%% file: main.tex
\documentclass[11pt]{article}

\usepackage[final]{acl}

\input{preamble/packages}

\input{preamble/definitions}

\input{preamble/authors}

\title{Stylization for Utility Enhancement in Annotation-Free Attribute-Based Synthetic Data Generation}
\title{Styles Does Matter in Annotation-Free Synthetic Data Generation}
\title{The Significance of Style Diversity in \\ Annotation-Free Synthetic Data Generation}

\begin{document}
\maketitle
\input{sections/abstract}

\input{sections/intro-4}

\input{sections/related2}

\input{sections/method}

\input{sections/experimentalsetup}
\input{sections/results}

\input{sections/conclusion}
\input{sections/limitations}
\input{sections/ethics}
\bibliography{references}

\appendix
\input{sections/appendix}

\end{document}

%% file: preamble/packages.tex
\usepackage{color}
\usepackage{flushend}
\usepackage{graphicx}
\usepackage{booktabs}
\usepackage[htt]{hyphenat}
\usepackage[]{xcolor}
\usepackage{tabularx}
\usepackage{multirow}
\usepackage[inline]{enumitem}
\usepackage{xspace}
\usepackage{acronym}
\usepackage{adjustbox}
\usepackage{subcaption}
\usepackage{arydshln}
\usepackage{amsmath} 
\usepackage{anyfontsize}
\usepackage{times}
\usepackage{latexsym}
\usepackage{pifont}
\usepackage[T1]{fontenc}
\usepackage[utf8]{inputenc}
\usepackage{microtype}
\usepackage{inconsolata}
\usepackage{makecell}

%% file: preamble/definitions.tex
\definecolor{midnightgreen}{rgb}{0.0, 0.29, 0.33}

\newcommand{\universalStylizer}{Univ\xspace}
\newcommand{\exampleStylizer}{Exam\xspace}
\newcommand{\appthree}{$\text{Univ}_{\text{StylePTB}}$\xspace}
\newcommand{\privateData}{EIC\xspace}
\newcommand{\llama}{Llama\xspace}
\newcommand{\roberta}{Dist-Roberta\xspace}

\acrodef{LLM}{Large Language Model}
\acrodef{SLM}{Small Language Model}
\acrodef{NLI}{Natural Language Inference}
\acrodef{IR}{Information Retrieval}
\acrodef{RAG}{Retrieval Augmented Generation}

\newcommand{\header}[1]{\vspace{2mm}\noindent\textbf{#1}}

%% file: preamble/authors.tex
\author{
  Zahra Abbasiantaeb$^1$ \quad 
  Zeno Belligoli$^2$ \quad 
  Omar Essam$^2$ \quad 
  Mohammad Aliannejadi$^1$ \\
  $^1$University of Amsterdam, Amsterdam, The Netherlands \\
  $^2$Booking.com, Amsterdam, The Netherlands \\
  \texttt{\{z.abbasiantaeb, m.aliannejadi\}@uva.nl} \\
  \texttt{\{zeno.belligoli, omar.essam\}@booking.com}
}

%% file: sections/abstract.tex
\begin{abstract}
Generating high-utility synthetic data for intent classification typically requires human-annotated seed data, which is often unavailable in fast-paced industrial settings. 
In this paper, we propose a framework for synthetic dialogue generation that works entirely without human-annotated data, relying solely on intent definitions.
Our proposed dialogue generation framework utilizes two different types of topic and style attributes to improve data diversity.
Also, we propose two novel post-hoc stylization models called \universalStylizer and \exampleStylizer to transform synthetic LLM-generated utterances into more varied, human-like linguistic styles.  
To enhance data quality, we utilize an LLM-as-a-judge filtering process. 
Experimental results on both industrial and public datasets demonstrate that the proposed approach achieves up to 93.3\% of the performance obtained using human-annotated training data. 
Crucially, the findings reveal that style diversity is more critical than topic diversity for synthetic data utility, as it prevents models from learning spurious stylistic correlations. 
Furthermore, the study shows that incorporating style attributes during the generation process is more effective than post-hoc style adaptation.
\end{abstract}

%% file: sections/intro-4.tex
\section{Introduction}

Using \acp{LLM} to generate synthetic data for augmentation with data collected by humans has shown improvement in different downstream tasks~\cite{Soudani2026}, including intent classification~\cite{mitra-etal-2026-recap}, dialogue state tracking~\cite{du-etal-2025-dflow}, and conversational recommendation~\cite{xu-etal-2025-beyond}.
Synthetic data can be generated using human-annotated data as few-shot examples~\cite{wang-etal-2023-self-instruct,schick-schutze-2021-generating} or in a zero-shot setting~\cite{AskariPMAAKV25,LinHV24} where no annotated data is used for generating the samples but is used to enhance the utility of the generated data.

\header{Annotation-free data generation.}
Although augmenting human-generated data is of high value, in various industrial scenarios, generating synthetic data without any prior human annotations (annotation-free) has many use cases.
Collecting high-quality, balanced data in such fast-changing environments is often unfeasible because human annotation is both time-consuming and expensive~\cite{ChenGKXL21}. Below, we list three potential scenarios where data generation with no prior human annotation is essential:
\begin{enumerate*}[leftmargin=*,nosep, label=\textbf{(S\arabic*)}]
    \item In a more extreme scenario, a company might be developing a completely new product and needs to develop and test it on synthetically generated data before launching it.
    \item As user needs evolve continuously, companies must adapt or update their existing models rapidly.     
    \item  Companies might need to transfer their existing models to new regions or for new applications, and would need to develop and test their models before they are put in production. 
    For instance, a travel platform in the US, launching a trip planner in India, would not have any trip-planning data from the Indian market. Each region has its own specific needs and characteristics, making it challenging to use and transfer the existing US-based data to India~\cite{DBLP:conf/emnlp/00020SA24}. Therefore, the company could leverage synthetic data to develop and test its new product or model under a fully simulated environment before its deployment.
\end{enumerate*}

\header{Linguistic style matters.}
A key challenge in annotation-free data generation is the lack of diversity. Using the same prompt without randomized few-shot examples (taken from human annotations) to generate thousands of samples could lead to highly repetitive outputs.
A common approach to introduce diversity is attribute-based generation~\cite{YuZZMRKSZ23,personahub,AskariPMAAKV25}, where specific details about the target data is provided via attributes (e.g., news topic, length, location, etc.).
Although recent research has shown that a lack of stylistic diversity can bias a model~\cite{Cao25}, existing synthetic data generation methods often fail to clearly distinguish between topic- and style-related attributes, overlooking the importance of style~\cite{YuZZMRKSZ23}. 
Current approaches typically represent style using superficial constraints, such as minimum or maximum word counts~\cite{YuZZMRKSZ23}, which LLMs often struggle to follow~\cite{Juncheng2025}, or broad phrasal descriptors (e.g., formal, informal, or aggressive) studied in the news article generation domain~\cite{YuZZMRKSZ23}. 
To the best of our knowledge, no prior work has systematically studied the impact and importance of linguistic style attributes in data generation, and its contribution to the utility of data. 

\header{Generation domain.}
With the rise of interest in task-oriented dialogue systems, building and adapting existing systems in new domains with limited to no available data is of great interest. Task-oriented dialog systems rely heavily on intent classification which involves categorizing a user's utterance into one of several pre-defined intents at every dialogue turn~\cite{NiYPXC23,HengstWAK24,KimYJ22,AlfieriWH22}. 
For example, the user utterance ``Can I change my flight?'' might be mapped to the ``change\_booking'' intent within a travel planner system. 
Typically, when a user initiates a new request, it is followed by several turns of dialogue between the user and the system~\cite{AlfieriWH22,PurverGH01}. 
These turns can be clarifying or elicitation questions, while still focusing on the same request.

\header{Contributions.}
In this work, we propose an annotation-free data generation framework for task-oriented dialogue systems.
We focus on studying how closely synthetic data generated without any human annotation can approach the performance achieved using human-annotated data in the intent classification task. 
Our proposed framework generates a chunk of dialogue for a given intent using intent definitions in one LLM call, reducing computational overhead.
We categorize attributes into two groups, namely, topic and style. Specifically, we define style attribute values that reflect user writing behavior when interacting with a dialogue system.
Through our experiments, we also study the importance of diversity within both topic and style attributes. 
Unlike existing work~\cite{AskariPMAAKV25,LinHV24}, to filter out the low-quality sample, we employ an \ac{LLM} as a judge instead of using existing human-annotated data.

In addition, we propose two stylization models, called \universalStylizer and \exampleStylizer, that transfer the style of \ac{LLM}-generated dialogues to match the human style of the target application domain. 
\universalStylizer learns to adapt the style of the input utterance into a universal human-like style while \exampleStylizer learns to adapt it to the style of the given examples in the input.
We compare our stylization models with existing style transfer datasets~\cite{StylePTBLyuLPHPSM21}.

We conduct experiments on both industrial and public datasets, achieving 90.7\% and 93.3\% of the accuracy obtained using human-annotated training data, respectively. 
Surprisingly, our findings demonstrate that style diversity is more important than topic diversity for the utility of synthetic data. 
Both \texttt{Llama-3.2-1B} and \texttt{distilroberta-base} benefit significantly from style diversity, whereas only \llama shows a small improvement from topic diversity.
We believe style diversity is more critical because it prevents the model from learning spurious correlations between stylistic features in the training data and intent classes. 
Our proposed stylization models improve the utility of synthetic data generated without any attributes by 4.7\%, demonstrating that adapting the style to the target domain enhances performance. 
Finally, we find that incorporating style attributes during the synthetic dialogue generation process proves more effective than generating data without style attributes and adapting the style afterward.\footnote{We will release the code upon acceptance.}

%% file: sections/related2.tex
\section{Related Work}

\header{Synthetic Dialogue Generation.} The use of \acp{LLM} for synthetic data generation has become a basis in addressing data scarcity. 
Recent work has moved beyond simple class-conditional generation toward attribute-based methods to mitigate inherent biases in LLM outputs. 
For instance, AttrPrompt~\cite{YuZZMRKSZ23} conditions generation on specific attributes like location and topic. 
Some of these attributes are class-dependent, while others remain class-independent~\cite{YuZZMRKSZ23}. For example, in news article generation, length and writing style serve as class-independent attributes, whereas the subtopic attribute depends on the specific article category, such as science, or economy~\cite{YuZZMRKSZ23}.
In addition, Persona Hub~\cite{personahub} leverages a massive scale of one billion distinct personas to capture diverse professional and personal perspectives.
Several frameworks have explored synthetic dialogue generation for intent classification and Dialogue State Tracking (DST), demonstrating performance gains when synthesized data is used for data augmentation~\cite{Mohapatra0CJ21,AskariPMAAKV25}.

The framework in~\cite{Mohapatra0CJ21} utilizes a small seed of original human dialogues and the specific instructions originally provided to crowd-workers to train its user and system simulators. Consequently, the synthetic data does not replace human data; rather, it serves as a method of augmenting a limited human-annotated foundation to improve model robustness in low-resource scenarios.
SOLID~\cite{AskariPMAAKV25} generates multi-turn dialogues in a zero-shot setting for intent classification by using entity-based seeds (e.g., name, occupation). 
However, SOLID and similar user-simulator frameworks~\cite{Mohapatra0CJ21} still require human-annotated data, for filtering low-quality dialogues and for borrowing intent sequences for dialog generation. 
Similarly, the Refiner framework proposed by~\cite{LinHV24} addresses zero-shot intent classification but is limited to single-turn utterances and relies on fine-tuning a refiner model on existing seen domains.
Also, while the Refiner learns to convert the style of \acp{LLM}-generated data to match human data, it does not provide explicit control over the semantic meaning of the generated outputs.
However, both the discussed models generate the samples in a zero-shot setting, but they still need human-annotated data.
For DST, research has focused on maintaining state consistency. 
\citet{FinchC24} addressed the narrowness of existing datasets by generating dialogues across over 1,000 diverse, automatically-derived domains. Their approach uses an LLM to first generate a specific scenario description and then simulate a conversation along with its corresponding silver-standard state annotations, ensuring the model is exposed to a wide array of linguistic contexts.
In contrast, SynthDST~\cite{KulkarniTMPYB24} prioritizes structural fidelity by using a schema-to-dialogue framework. Instead of free-form generation, they use utterance-level prompting anchored to predefined dialogue templates and schema. This ensures that the generated text remains strictly grounded in the underlying slot-value pairs, effectively mimicking the quality of human-annotated few-shot data.
Unlike these approaches, which often rely on predefined templates or external annotated data, our proposed framework is entirely independent of human-annotated data.

\header{Style transfer and diversity.} Recent literature emphasizes that stylistic uniformity in training data can introduce systematic biases. For example, Information Retrieval (IR) models have been shown to favor formal or academic prose, leading to unfair outcomes in ranking~\cite{Cao25}. While benchmarks like StylePTB~\cite{StylePTBLyuLPHPSM21} provide datasets for fine-grained, compositional style transfer (e.g., altering tense or voice), they often focus on atomic linguistic changes rather than the complex behavior of real-world user interactions.
Our work bridges this gap by proposing two novel style transfer models. 
Unlike existing benchmarks, our approach establishes a direct mapping between real user behavior and LLM-generated outputs.

%% file: sections/method.tex
\input{tables/attribute}

\section{Methodology}
Intent classification in task-oriented dialogue systems is defined as classifying the user's utterance in every turn of the dialogue into one of the existing intents. 
Each dialogue  $D^{(l)}  = \{(u_1, r_1),..., (u_l,r_l)\}$ includes a set of user ($u_i$) and system ($r_i$) utterances where ($l$) is the length of the dialogue. 
The intent classification function is defined as classifying each user utterance $u_i$ into one of the intent classes of $C=\{c_1,\dots,c_m\}$.

\subsection{Synthetic Dialogue Generation}
\looseness=-1
We address the problem of annotation-free synthetic dialogues generation for intent classification using only intent definitions.
Our dialogue generation model takes a sequence of intents as input and generates a chunk of dialogue for each intent.
Each chunk includes multiple turns (1--5 turns) of user--system utterances, starting with the user utterance asking a question with the given intent. 
The system can respond to the user's question or ask clarifying or elicitation questions. 
As a dialogue evolves, the user continues asking questions relevant to the intent or responding to system questions.
This is in line with the definition of intent in existing datasets and applications, where the intent is maintained during elicitation and clarification. 
Different from existing work~\cite{AskariPMAAKV25}, which borrows the possible sequences of intents from existing datasets to generate realistic and natural dialogues, we use prior knowledge of a strong LLM to generate different possible sequences of intents by zero-shot prompting (See Table~\ref{tbl:prompts-sequence-gen} in Appendix).
The designed prompt leverages the definition of the intents (see Table~\ref{tbl:prompts-dialog_gen} in Appendix~\ref{sec:app-prompts} for the prompt).
An example of a class definition is shown in Table~\ref{tbl:class-def} in Appendix.
A random sample of the generated sequences is later used for dialogue generation.
We define our dialogue generation function ($F$) as follows: 
\begin{equation}
\label{eq:dialog_gen}
    \tilde{D}^{(i)} = F(c_i,\, t,\, w,\, D^{(i-1)}), \text{ for each } c_i \in S~,
\end{equation}
\begin{equation}
    D^{(i)} \leftarrow D^{(i-1)} \oplus \tilde{D}^{(i)}~,
\end{equation}
\noindent where $S$ is a sequence of intents, $c$ is a class of intent, $w$ and $t$ are the style topic attributes.
Each class $c$ is represented by its name and definition. 
Each chunk of chat $\tilde{D}^{(i)}$ includes 1--5 turns of user--system utterances and $D^{(i)}$ represents the full dialogue generated till step $i$.
An example workflow of our dialogue generation function is shown in Figure~\ref{fig:gen-framework} in Appendix~\ref{sec:datagen-framework}.

\header{Topic attribution.}
The topic attributes are fine-grained information about the topic of the conversation and the user. Attributes can be either class-dependent or class-independent~\cite{YuZZMRKSZ23}.
Examples of class-independent attributes are destination country, number of travelers, and number of adults. 
An example of intent-dependent attributes for the intent class of customer service can be the type of the user request or complaint.
More examples of attributes are provided in Table~\ref{tab:attribute-values} in the Appendix.
Since we assume no access to annotated data, we manually define the attribute dimensions based on domain knowledge and high-level business requirements. We then generate a list of possible values for each attribute dimension using zero-shot prompting the \acp{LLM} (see Table~\ref{tbl:prompts-attr-gen} in Appendix for prompt).
The attribute dimensions can also be generated with the assistance of an LLM by first prompting the model to propose a set of possible dimensions, after which a subset of relevant dimensions is selected.

\header{Style attribution.}
The style attribute defines the user's writing style in the dialogue. 
These attributes are tailored to capture the user’s communication behavior with the chatbot and include aspects such as text formatting (e.g., use of lowercase or uppercase letters, emojis, and punctuation) and language type (e.g., formal language that is polite and grammatically correct).
We define a set of style attributes with the aid of LLM.
First, we prompt the LLM to generate a set of user behavior styles when interacting with dialogue systems, and then we manually inspect and refine them.
The defined style attributes align with the definitions proposed by~\cite{Cao25}, but are tailored specifically to capture user linguistic behavior within the context of dialogue system interactions.
We provide examples of these attributes in Table~\ref{tab:attributes}.

\header{Filtering.}
As LLMs are not perfect, some of the turns generated by them might not have the given intent.
To identify such cases, we use another \ac{LLM} to detect the intent of each user utterance and we do not use the turn as a training sample if the intent predicted by the other \ac{LLM} is different from the original intent. 
The prompt designed for this task includes the intent names and definitions with one example demonstration for each intent (see Table~\ref{tbl:prompts-filtering} in Appendix for prompt).

\subsection{Stylization}
To diversify the style of the user utterance generated by \ac{LLM}, we rely on two different approaches, namely, (i) in-context style attributes and (ii) post-stylization. %
In the in-context stylization approach, we include the user's writing style $w$ in the prompt, as an attribute. 
In the post-stylization approach, on the other hand, we first prompt the LLM to generate the utterance, and then run a stylization model.

\header{Post-stylization.}
We propose two different post-stylization approaches, namely \universalStylizer and \exampleStylizer.

\begin{itemize}[leftmargin=*,nosep]
    \item \textbf{Universal (\universalStylizer)} learns a general human-like linguistic style and converts the style of the text generated by \ac{LLM} into a text similar to human (See Figure~\ref{fig:stylization}).
    The \universalStylizer stylization function is defined as follows: \looseness=-1
    \begin{equation}
        \tilde{u}_i = \text{StylizeUni}(r_i, u_i)~.
    \end{equation}
    As the user utterance depends on the previous system response, the input of the model includes the previous system utterance ($r_i$) and the current \ac{LLM} utterance ($u_i$). The output of the model is the stylized user utterance ($\tilde{u}_i $).
    To train the universal model, we need a training dataset that maps the \ac{LLM}-utterance into a human-written utterance.
    To collect such data, we use existing dialogue datasets and pass the dialogue up to turn $n$ to an \ac{LLM} and instruct it to respond to the last system utterance ($r_n$) with the same content of the last user utterance ($\tilde{u}_n$) using its own linguistic style (see prompts in Tables~\ref{tbl:prompts-style-data-gen} and~\ref{tbl:prompts-style-univ} in Appendix). 
    We use the utterance generated by LLM as $u_i$ in the above function.
    We instruct the \ac{LLM} to generate 5 different sentences for each user utterance.
    We assume that this transition preserves the meaning, content, and intent of the user.
    Hence,  have a mapping between a real user utterance and a synthetic utterance with \ac{LLM}.
    We use the \ac{LLM}-generated utterance as ($u_i$) and the original human utterance from the existing dialogue as ($\tilde{u}_i $) for training the above function.  \looseness=-1
    \item \textbf{Example-based (\exampleStylizer)}: 
    The universal stylization is effective in transferring the style, but can be limited in generating a diverse set of styles. It is trained to map synthetic datasets with the same style (as they are generated by the same \ac{LLM}) to human utterances with different styles (as humans are different in this setting).
    To mitigate this limitation, we include some example utterances of the user in the input to show examples with the same linguistic style as the stylized utterance to the model (see Figure~\ref{fig:stylization}).
    Hence, the model will learn to change the style of the LLM utterance according to the human utterances provided in the context. 
    We define the example-based stylization function as follows:
    \begin{equation}
        \vspace{-1mm}
        \tilde{u}_i = \text{StylizeExam}(h ,r_i, u_i)~,
        \vspace{-1mm}
    \end{equation}
    where $h = \{(r_1, \tilde{u}_1),..\}$ is a set of several system and user utterances from the same user.
    For training, we use the same data collected for the universal model (see prompt in Table~\ref{tbl:prompts-style-examp} in Appendix~\ref{sec:app-prompts}). 
    The only difference is that we put several turns of user--system from the same user as $h$ in the input of the model.
\end{itemize}

%% file: tables/attribute.tex
\begin{table}[]
    \centering
    \small
    \caption{Examples of user writing style attribute values.}
    \resizebox{\columnwidth}{!}{
    \begin{tabular}{p{7.3cm}}
    \toprule
       \textbf{Formal question}: User uses formal language for questions which is polite and grammatically correct. User ends sentences without punctuation. \\ \midrule
       \textbf{Direct request}: User uses direct, straightforward requests, often in imperative or declarative form and does not repeat the name of the named entities when mentioned previously.   \\ \midrule
       \textbf{Short keyword-style query}: User uses short keyword-style sentences when asking questions and responds to clarification or elicitation questions with minimal words.\\ \midrule
       \textbf{Command}: User gives direct commands, often starting with a verb, expecting immediate action. \\ \midrule
       \textbf{Colloquial/slang}: User uses colloquial language or slang, including contractions or regional expressions. User sometimes uses emojis.\\ \midrule
       \textbf{Aggressive}: User uses aggressive language, often demanding or forceful in their requests. \\
       \bottomrule
    \end{tabular}
    }
    
    \label{tab:attributes}
\end{table}

%% file: sections/experimentalsetup.tex
\section{Experimental Setup}
\header{Datasets.}
To assess the effectiveness of our proposed dialogue generation framework for intent classification, we generate two synthetic training datasets based on the Schema-Guided Dialogue (SGD) dataset and an anonymized proprietary industrial dataset called Enterprise Intent Corpus (\privateData).
The SGD~\cite{SGDRastogiZSGK20} is a public large-scale dataset designed for task-oriented dialogue systems, covering over 20 domains and 40 services.
In this dataset, each user utterance is annotated with exactly one intent.
We use the description of intents provided in the schema of the dataset as class definitions.
As we need to compare our method to the scenario of only using human-annotated data, we select a subset of common intents in both the train and test subsets for synthetic data generation. 
The  \privateData is collected from a real-world task-oriented trip planning dialogue system. 
The statistics of the datasets are provided in Table~\ref{tab:dataset-stat} in Appendix~\ref{sec:app:dataset}. 
As we can see, the \privateData dataset is much smaller than SGD dataset. In \privateData dataset, we identify 4 minority classes with lower than 35 samples in training set while more than 50\% of the sample belong to a single class.

\header{Stylization datasets.} The synthetic dataset collected for \exampleStylizer and \universalStylizer stylization models across \privateData includes 23,011 / 4,826 / 4,611 samples for train/test/validation splits, respectively.
Also, for the SGD dataset, we use a total of 39,931 / 9,261 / 8,771 samples for train/test/validation splits, respectively.

\header{Synthetic datasets.} Using our framework, we generate synthetic training sets for the SGD and \privateData datasets. For SGD, we generate 3,000 synthetic dialogues comprising 36,188 turns. For \privateData, we generate 11,017 synthetic dialogues comprising 63,945 turns. After filtering, 2,451 and 5,743 turns are removed from the SGD and \privateData datasets, respectively.

\header{Baselines.} To assess the effectiveness of our proposed models, we compare them to the following baselines.
\begin{itemize}[leftmargin=*,nosep]
    \item No-Attribute generation: A naïve baseline where dialogues are generated in a zero-shot manner without specifying any stylistic or topical constraints, only using class definitions. For this baseline method, we use the prompt designed for our framework and remove the topic and style attributes from it.
    \item Topic-only generation: Inspired by recent works in attribute-guided data expansion~\cite{Li2025} and attributed training data generators~\cite{AskariPMAAKV25,YuZZMRKSZ23}, this baseline utilizes only topic attributes (e.g., domain, intent) while omitting stylistic attributes. To implement this baseline, we omit the style attribute from the prompt designed for our method.
    \item \appthree: To compare the quality of our collected dataset with existing style transfer datasets, we train a version of the \universalStylizer stylization using the StylePTB dataset~\cite{StylePTBLyuLPHPSM21}. We generate synthetic dialogues using the ``No-Attribute generation'' baseline and then transfer its style using this stylization model.
    \item Human: We include the original human-labeled training set as a performance upper bound to measure the utility gap between synthetic and real-world data distributions.
\end{itemize}

\header{Evaluation metrics.} 
We explain the metrics used for evaluation in Appendix~\ref{sec:metrics}.

\header{Models and parameters.} We report the language models we use for dialogue generation, stylization, and intent classification, as well as their corresponding hyper-parameters in Appendix \ref{sec:params}.

%% file: sections/results.tex
\section{Results}

\input{tables/main-results}

\header{Synthetic data vs human data.} 
As shown in Table~\ref{tab:downstream-perf}, when using the \llama (\roberta) model for intent classification, our synthetic dataset achieves 90.7\% (86.3\%) and 93.3\% (89.6\%) of the accuracy obtained with human-annotated data on SGD and \privateData, respectively.
Notably, when utilizing the \llama model on the \privateData dataset, the synthetic data actually yields a higher F1 score than the human-annotated data. 
This improvement is due to the presence of minority classes within the dataset, suggesting that our approach is particularly effective at enhancing model performance for underrepresented classes.

\input{tables/intent-detection-llms}
\header{Impact of filtering.}
We evaluate various LLMs for intent classification on the test set of the \privateData dataset, with results summarized in Table~\ref{tab:llm-filtering}. 
Among the tested models, \texttt{gpt-4-1} achieves the highest performance. 
Consequently, we use this model for filtering. 
As shown in Table~\ref{tab:ablation}, filtering these inconsistent instances significantly enhanced the dataset's utility, leading to improved performance in the downstream task.

\input{tables/ablation-attributes}

\header{Impact of style and topic attributes.}
To evaluate the individual contributions of style and topic attributes, we compare our proposed method against No-attribute and Topic-only baselines. 
For a fair comparison, we use the same intent sequences and attributes to generate 1,000 synthetic dialogues for Ours method and baselines.
As shown in Table~\ref{tab:ablation}, the style attributes used in our approach are the primary drivers of performance. 
The G-Vendi score~\cite{jung2026prismatic} has been shown to be an effective proxy metric for evaluating the utility of synthetic data in downstream tasks. 
We use the \roberta model to compute G-Vendi scores for the different synthetic datasets generated for the \privateData dataset. 
The Ours, Ours (Style-only), No-attribute, and Topic-only datasets has G-Vendi scores of 9.54, 9.46, 9.21, and 9.36, respectively. 
As shown, the dataset generated by our method (which demonstrates the best performance on the downstream task) also yields the highest G-Vendi score.
This emphasis on style aligns with existing research~\cite{Cao25} suggesting that LLMs often associate superficial stylistic features with specific classes rather than mastering underlying semantic concepts. 
Our findings suggest that increasing style diversity of the synthetic data prevents models from learning these short-circuit mappings. 
In contrast, topic attributes (such as travel destination or facility preferences) primarily introduce diversity in named entities and numerical values.
On the SGD dataset, utilizing only style yields the best results. 
On the \privateData dataset, the addition of topic attributes provides  marginal gains only when using the \roberta model.
Comparing the Topic-only and No-attribute baselines, we observe that topic diversity improves utility for the \llama model but actually degrades performance for \roberta. 
This suggests that while LLMs may benefit from topical variety, \acp{SLM} may not benefit from it. This notable finding challenges the conventional emphasis on increasing topical diversity to improve downstream performance.

\input{tables/stat}

Moreover, comparing No-attribute and Ours (Style-only) models, we observe that removing the style attribute has a more negative impact on the \privateData dataset than on the SGD dataset. 
This discrepancy highlights a fundamental difference between academic and real-world data. 
While public datasets like SGD are typically curated by a limited number of trained crowd workers, \privateData dataset originates from a production environment. 
Consequently, it captures a broader spectrum of linguistic diversity, reflecting users from various regions, languages, and educational backgrounds. 
These findings suggest that style diversity is even more critical in real-world applications where user input is less standardized than in academic benchmarks.
The linguistic analysis in Table~\ref{tab:linguistic} reveals clear stylistic differences between the two datasets. 
Higher values for TTR and the Hapax Ratio show that the \privateData dataset has much greater vocabulary diversity than SGD. 
This is further supported by a higher Entropy score, which indicates that the language in \privateData is more varied and less predictable.
Readability scores (Gunning Fog index and Flesch Reading Ease) also highlight a contrast in tone: the \privateData dataset uses a more complex and technical register, whereas SGD follows a simpler, more conversational structure. 
Furthermore, while the average sentence length is shorter in \privateData, a higher standard deviation reveals an irregular flow, contrasting with the more uniform and structurally deep sentences found in SGD.
\input{tables/stylizaers-results}

To further study stylistic bias, we compare two setups: an all-synthetic setting, where all intent classes use synthetic training data, and a mixed-data setting, where only two target classes remain synthetic and rest of the classes use equally sized human data.
Our hypothesis is that if the model is truly learning semantic representations, its performance on the target classes should remain consistent across both setups. 
Surprisingly, we observe a significant performance collapse on the synthetic-only classes in the mixed-data setting compared to the all-synthetic.
The average of F1 metric for these classes drops from 0.84 to 0.38.
This finding suggests that in the presence of real human data, the model identifies the consistent stylistic signature of the LLM as a discriminative feature for the target classes. 
However, when stylistic diversity is introduced or when the entire dataset shares the same synthetic origin as test data, the model is forced to move beyond these surface-level patterns and focus on the actual meaning of the utterances.

\input{tables/compare-stylizers}

\header{Impact of stylization models.} 
We train both T5 and \llama-based variants of our proposed stylization models and apply them over the data generated by No-attribute baseline.
As shown in Table~\ref{tab:stylizer}, our proposed stylization models (\universalStylizer and \exampleStylizer) improve the utility of synthetic data for the \llama model compared to the No-attribute baseline. 
Notably, these models outperform the \appthree baseline. 
While the StylePTB dataset focuses on surface-level changes (such as synonym replacement, tense shifts, or the addition of modifiers) our results suggest that these transformations are insufficient. 
However, we observe while stylization improves performance for \llama, it does not provide similar gains for \roberta. 
Our proposed method of  using style attributes during generation, is more effective than post-stylization methods in all cases except using \llama for intent classification across SGD dataset. 
This confirms that defining and integrating user writing styles during the initial generation phase is the most effective way to produce diverse, high-utility dialogues.

\header{Evaluating stylization quality.}
We further assess our stylization models using BLEU and accuracy metrics (Table~\ref{tab:stylizer-evaluate}). 
A key observation is that stylization can occasionally compromise user intent. 
For instance, the T5-based \universalStylizer altered the original intent in 7.4l\% of cases, according to the intent classifier LLM we use for filtering.
Despite this trade-off, stylization significantly increased the lexical similarity between LLM and human utterances while T5 increases the lexical similarity more than \llama.
Also, the \exampleStylizer achieves higher BLEU compared to the \universalStylizer.
Interestingly, the BLEU similarity between LLM-generated utterances and the human utterances from SGD dataset (7.4) is higher than that of the \privateData (4.13), suggesting that standard LLM outputs are naturally more aligned with curated public datasets. 
Following stylization, the similarity to \privateData improved by 16.67 points, nearly double the 9.7 point improvement seen in the SGD dataset. 
This disparity highlights a critical insight. 
The public datasets, often produced by trained annotators, possess lower syntactic diversity and are closer to standard LLM outputs. 
In contrast, real-world industry datasets exhibit much higher style diversity, making stylization an indispensable step for synthetic data generation in production environments.
Finally, we evaluate intent classification performance after applying Filtering (Table~\ref{tab:stylizer-filtered}). 
To ensure a fair comparison, we utilized a union-filtered subset, where a dialogue turn was removed across all methods if it was filtered in any single method. Even with the filtering, we confirm that stylization consistently improves data utility for the \llama model, whereas the \roberta model remains less sensitive to these stylistic enhancements across \privateData dataset.

\input{tables/stylization-filtered}

%% file: tables/main-results.tex
\begin{table}[]
    \centering
    \small
   \caption{Intent classification performance on the \privateData and SGD datasets using synthetic (Synth) and real datasets. Metrics include Accuracy (Acc) and Macro F1 (F1). The model used for intent classification is shown with (IC). Filtering is applied over the synthetic data.}
\resizebox{\columnwidth}{!}{    
\begin{tabular}{llccccc}
\toprule
 \multirow{2}{*}{\textbf{Dataset}}       &   \multirow{2}{*}{\textbf{IC}}   & \multicolumn{2}{c}{\textbf{\privateData}}  &   \multicolumn{2}{c}{\textbf{SGD}}   \\ \cmidrule{3-6}
 
& & \textbf{F1} &  \textbf{Acc}  &  \textbf{F1} &  \textbf{Acc}  \\ \toprule
Human   & \multirow{2}{*}{\rotatebox{0}{\llama}}  & 0.728  & 0.843 & 0.863& 0.875\\
Synth  &  & 0.788  & 0.765 & {0.777} & {0.817}\\  \midrule

Human   & \multirow{2}{*}{\rotatebox{0}{\roberta}}   & 0.791  & 0.851 & 0.925 & 0.928 \\
Synth  & & 0.748  & 0.735 & 0.801 &  0.832\\ 
 \bottomrule
    \end{tabular}
    }
    \label{tab:downstream-perf}
\end{table}

%% file: tables/intent-detection-llms.tex
\begin{table}[]
\small
 \centering
 \caption{Intent classification performance of different LLMs. The best results are in \textbf{bold}.}
 \resizebox{\columnwidth}{!}{
  \begin{tabular}{lcccc}
   \toprule
  \textbf{LLM} & \textbf{Precision} & \textbf{Recall} & \textbf{F1} & \textbf{Kappa}  \\ \toprule
  \texttt{gpt-4-1-mini}& 0.478 & 0.731 & 0.535 & 0.649 \\  
  \texttt{gpt-4-1}     & \textbf{0.579} & \textbf{0.842} & \textbf{0.642} & \textbf{0.709} \\  
  \texttt{gpt-5-mini}  & 0.501 & 0.679 & 0.523 & 0.471 \\ 
  \bottomrule 
 \end{tabular}
 }
 \label{tab:llm-filtering}
 \end{table}

%% file: tables/ablation-attributes.tex
\begin{table}[]
\small
\caption{Performance of intent classification using synthetic data generated by our method and baselines.  Filtering is shown with (F).}
\label{tab:ablation}
\resizebox{\columnwidth}{!}{
\begin{tabular}{llcccccc}
\toprule

\multirow{2}{*}{\textbf{IC}}& \multirow{2}{*}{\textbf{Model}}  & \multicolumn{2}{c}{\textbf{\privateData}} & \multicolumn{2}{c}{\textbf{SGD}} \\ \cmidrule{3-6}
& &  \textbf{F1}  & \textbf{Acc} &  \textbf{F1}  & \textbf{Acc} \\
\toprule

\multirow{5}{*}{\rotatebox{90}{\llama}}
& No-attribute    & 0.706 &  0.660  & 0.727  & 0.748\\ 
& No-attribute (F) & 0.725 &  0.688 &  0.745 & 0.768 \\  
& Topic-only       & 0.743 &  0.715 & 0.736  & 0.766 \\
& Ours              & 0.735 &  0.722 & 0.736  & 0.774\\
& Ours (Style-only) & \textbf{0.763} &  \textbf{0.751} & \textbf{0.746}  & \textbf{0.781} \\
\midrule
\multirow{5}{*}{\rotatebox{90}{\roberta}}
& No-attribute       & 0.652 &  0.634 & 0.760 & 0.795\\ 
& No-attribute (F)   & 0.664 &  0.661  &  0.774 & 0.808 \\ 
& Topic-only         & 0.624 &  0.595 & 0.734 & 0.764 \\
& Ours                & \textbf{0.685} &  \textbf{0.671} & 0.752 & 0.785\\
& Ours (Style-only)   & 0.678 &  0.662 & \textbf{0.781} & \textbf{0.813}\\
\bottomrule
\end{tabular}
}
\end{table}

%% file: tables/stat.tex
\begin{table}[]
\centering
\small
\caption{Comparison of linguistic metrics between \privateData and SGD datasets.}
\label{tab:linguistic}
\resizebox{0.65\columnwidth}{!}{\begin{tabular}{lcc}
\toprule
\textbf{Metric} & \textbf{\privateData} & \textbf{SGD} \\
\toprule
TTR & 7.24 & 2.73 \\
Entropy & 8.84 & 7.72 \\
Flesch Reading Ease & 48.1 & 92.9 \\
Gunning Fog & 17.3 & 4.6 \\
Hapax Ratio & 3.5 & 1.12 \\
Avg. Sentence Length & 5.77 & 8.09 \\
Std. Sentence Length & 6.33 & 5.00 \\
Avg. Tree Depth & 2.28 & 2.89 \\
\bottomrule
\end{tabular}}
\end{table}

%% file: tables/stylizaers-results.tex
\begin{table}[]
\small
\caption{Comparison of intent classification performance across stylization models \textbf{without Filtering}. The backbone LLM used for training the stylization model is shown inside the bracket.}
\label{tab:stylizer}
\resizebox{\columnwidth}{!}{\begin{tabular}{llcccccc}
\toprule
\multirow{2}{*}{\textbf{IC}} &\multirow{2}{*}{\textbf{Stylization}} & \multicolumn{2}{c}{\textbf{\privateData}} & \multicolumn{2}{c}{\textbf{SGD}} \\ \cmidrule{3-6}
& &   \textbf{F1} & \textbf{Acc} & \textbf{F1} & \textbf{Acc} \\
\toprule

\multirow{9}{*}{\rotatebox{90}{\llama}}
 & No-attribute & 0.706 & 0.660 & 0.727 & 0.748 \\ 
 & Ours          & \textbf{0.763} & \textbf{0.751} & 0.746 & 0.781 \\
\cmidrule{2-8}
 & \universalStylizer [T5] & 0.702 & 0.707 & 0.725 & 0.751 \\
 & \exampleStylizer[T5]   & 0.684 & 0.690 & \textbf{0.766} & \textbf{0.790} \\ 
 & \appthree [T5]         & 0.718 & 0.736 & 0.752 & 0.760 \\
\cmidrule{2-7}
 & \universalStylizer [\llama] &  0.702 & 0.712 & 0.747 & 0.771 \\ 
 & \exampleStylizer   [\llama] &  0.700& 0.641 & 0.706 & 0.741 \\
 & \appthree [\llama]          & 0.607 & 0.624 & 0.718 & 0.736 \\
 
\midrule
\multirow{9}{*}{\rotatebox{90}{\roberta}}
 & No-attribute  & 0.652 & 0.634 & 0.760 & 0.795 \\ 
 & Ours           & \textbf{0.678} & \textbf{0.662} & \textbf{0.781} & \textbf{0.813}\\
\cmidrule{2-8}
 & \universalStylizer  [T5] & 0.610 & 0.604 & 0.767 & 0.804 \\

 & \exampleStylizer  [T5]  & 0.608 & 0.630 & 0.773 & 0.809 \\
 & \appthree [T5]          & 0.634 & 0.643 & 0.755 & 0.792 \\
\cmidrule{2-8}
 & \universalStylizer [\llama] & 0.617& 0.612 & 0.761 & 0.797 \\

 & \exampleStylizer  [\llama] & 0.619 & 0.640 & 0.729 & 0.763 \\
 & \appthree [\llama]& 0.612 & 0.631 & 0.759 & 0.791 \\

\bottomrule
\end{tabular}}
\end{table}

%% file: tables/compare-stylizers.tex
\begin{table}[]
    \centering
    \small
    \caption{Comparison of stylization quality evaluating the preservation of semantic intent (Corr) and the linguistic similarity (BLEU) between stylized outputs and reference texts across the test set. }
    \begin{tabular}{lcccc}
    \toprule
\multirow{2}{*}{\textbf{Stylization}} & \multicolumn{2}{c}{\textbf{\privateData}}&  \multicolumn{2}{c}{\textbf{SGD}} \\ \cmidrule{2-5}
     &  \textbf{BLEU}    & \textbf{Corr} & \textbf{BLEU}    & \textbf{Corr} \\
    \toprule
    No-attribute                & 4.13 & \textbf{97.8} &  7.4  & \textbf{94.7} \\ \midrule
    \universalStylizer [T5]     & 20.2 & 91.3 & 17.1  & 94.4 \\ 
    \exampleStylizer   [T5]     & \textbf{22.8} & 90.4 & \textbf{19.0}  & 93.8 \\ \midrule
    \universalStylizer [\llama] & 10.3 & 91.1 & 10.6  & 94.1 \\ 
    \exampleStylizer   [\llama] & 13.4 & 93.9 & 10.9  & 86.5 \\
 
    \bottomrule
    \end{tabular}
    \label{tab:stylizer-evaluate}
\end{table}

%% file: tables/stylization-filtered.tex
\begin{table}[]
\small
\caption{Comparison of intent classification performance using different stylization models \textbf{with Filtering}.}
\label{tab:stylizer-filtered}
\begin{tabular}{llcccc}
\toprule
\multirow{2}{*}{\textbf{IC}}&\multirow{2}{*}{\textbf{Stylization}}  & \multicolumn{2}{c}{\textbf{\privateData}}  & \multicolumn{2}{c}{\textbf{SGD}} \\ \cmidrule{3-6}
& &  \textbf{F1} & \textbf{Acc} & \textbf{F1} & \textbf{Acc} \\
\toprule

\multirow{3}{*}{\rotatebox{90}{\llama}}
 & No-attribute            & 0.725 & 0.688 & 0.745 & 0.768\\ 
\cmidrule{2-6}
 & \universalStylizer [T5] & 0.719 & \textbf{0.726} &  0.740 & 0.768\\
 & \exampleStylizer  [T5]  & \textbf{0.727} & 0.714  & \textbf{0.767} & \textbf{0.800}\\

\midrule
\multirow{3}{*}{\rotatebox{90}{\roberta}}
\\
 & No-attribute            & 0.664 & \textbf{0.661} & 0.774 & 0.808 \\ 
 \cmidrule{2-6}
 & \universalStylizer [T5] & \textbf{0.665} & 0.648 & 0.789 & 0.824\\
 & \exampleStylizer  [T5]  & 0.657 &0.656  & \textbf{0.795} & \textbf{0.831}\\ 

\bottomrule
\end{tabular}
\end{table}

%% file: sections/conclusion.tex
\section{Conclusion}
This paper presents a annotation-free framework for synthetic dialogue generation that eliminates the need for human intervention by only utilizing intent definitions and automated stylization. 
Our results on both industrial and public datasets demonstrate that synthetic data can achieve over 90\% of the performance of human-annotated data in intent classification tasks. 
Our experiments reveal that style diversity is a more critical factor than topic diversity in enhancing the utility of synthetic data. 
By incorporating varied user linguistic behaviors (either through attribute-based generation or our novel stylization models) we effectively mitigate the risk of models learning spurious stylistic correlations. 
Furthermore, we find that integrating style attributes directly into the generation process is superior to post-hoc stylization. 
Future work will explore the extensibility of this framework to more complex conversational tasks beyond intent classification, further reducing the reliance on costly human annotation in rapid deployment cycles.

%% file: sections/limitations.tex
\section{Limitations}
In this study, we conducted experiments on the intent classification task using only \roberta and \llama models.
For synthetic data generation, we used only the GPT-4.1-mini model and did not evaluate other LLMs.
In addition, our findings regarding the importance of style diversity in the utility of synthetic data are limited to the intent classification task. We leave the evaluation of other LLMs for data generation, the use of additional intent classification models, and the study of style diversity in other applications as future work.

%% file: sections/ethics.tex
\section{Ethical Considerations}
The authors identify and address several ethical and social implications related to the development of synthetic data for dialogue systems:
\begin{itemize}[leftmargin=*,nosep]
    \item Mitigating Bias: By diversifying linguistic styles, our framework prevents models from favoring specific prose types (e.g., formal vs. colloquial), ensuring fairer outcomes across varied user populations.
    \item Data Privacy: Generating training data from definitions rather than real user logs enables model development in sensitive domains without compromising user privacy.  
    \item Semantic Integrity: Our stylization models are specifically designed to transform linguistic expression while strictly preserving the original intent and meaning.  
    \item  Supporting Equity: The framework specifically improves performance for underrepresented minority classes, which are typically disadvantaged by a lack of human-annotated data.
\end{itemize}

%% file: sections/appendix.tex
\section{Appendix}
\begin{figure*}
    \centering
    \includegraphics[width=0.99\linewidth]{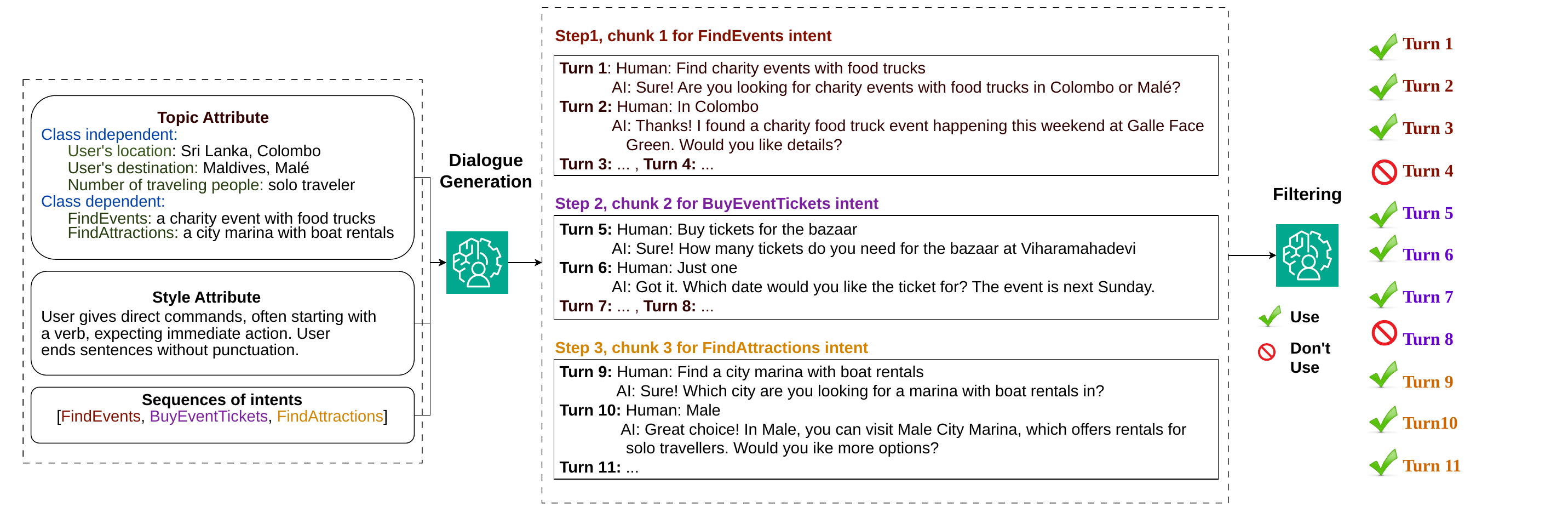}
    \caption{An example workflow of our dialogue generation model. The topic attributes, style attribute and sequence of intents are randomly selected from an existing pool.}
    \label{fig:gen-framework}
\end{figure*}

\subsection{Data Generation Framework}\label{sec:datagen-framework}
\begin{figure}
    \centering
    \includegraphics[width=0.9\linewidth]{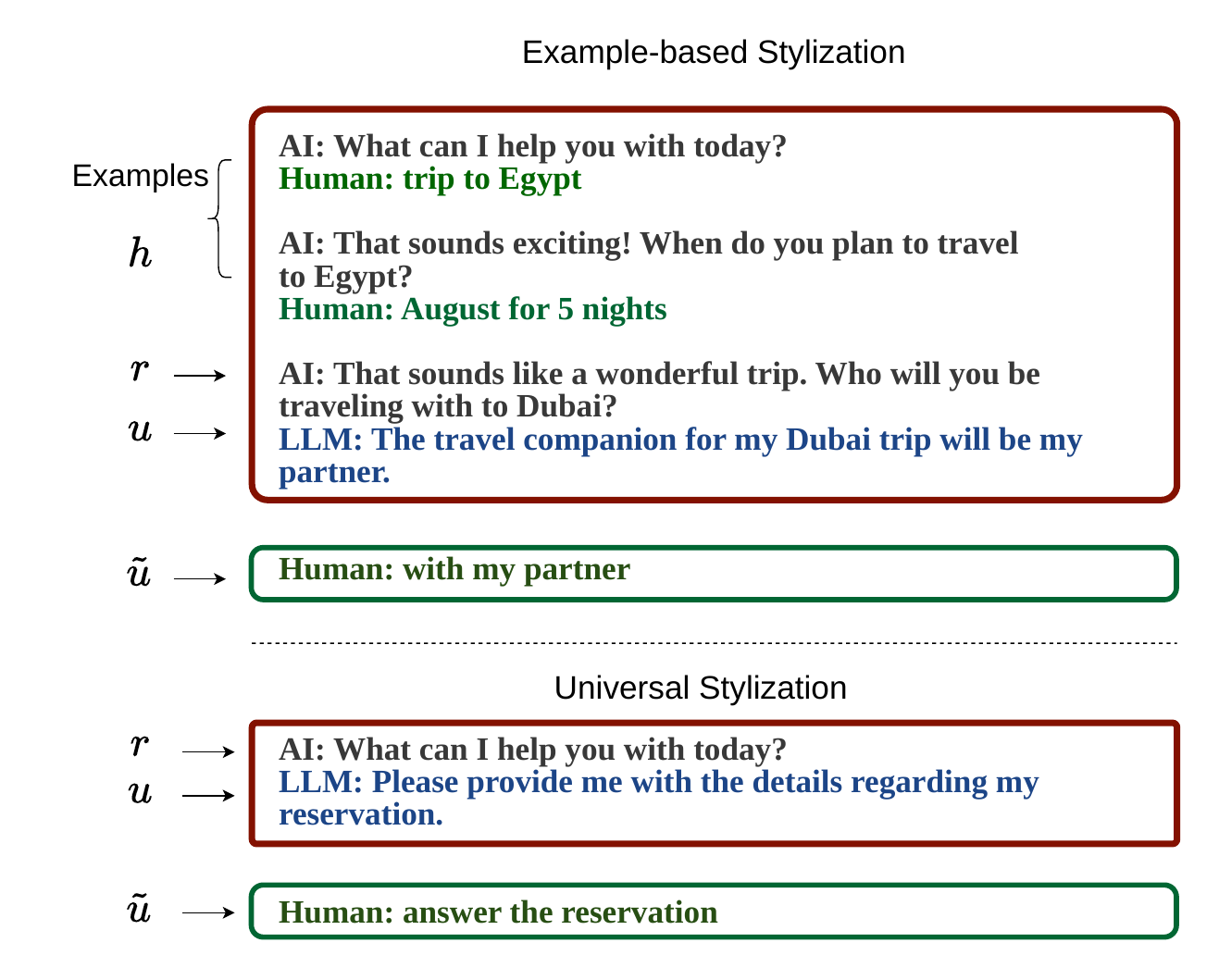}
    \caption{Two different proposed stylization models.}
    \label{fig:stylization}
\end{figure}

In Figure~\ref{fig:gen-framework}, we illustrate an example workflow of our data generation framework. As we can see, the dialogue generation LLM, generates a chunk of dialogue for each intent in a single LLM call. After generation, the Filtering LLM identifies the intent of each turn. If the intent predicted by filtering LLM is different from the original intent, the corresponding turn is not used as an training sample.
An example workflow of the proposed stylization models (i.e., \universalStylizer and \exampleStylizer) is shown in Figure~\ref{fig:stylization}.
The synthetically generated user utterance in dialogue generation framework (without using style and topic attributes) is passed as $u$ to these stylization models.

\subsection{Statistics of the Datasets.}\label{sec:app:dataset}
We report the statistics of the SGD and the \privateData datasets in Table~\ref{tab:dataset-stat}.

\subsection{Evaluation Metrics.}\label{sec:metrics}
 We evaluate the performance of intent classification models in terms of accuracy and macro F1. 
To provide a comprehensive view of the stylistic differences, we employed several measures of lexical diversity and structural complexity. 
Type-Token Ratio (TTR) and the Hapax Ratio~\cite{ali2014comparative} serve as primary indicators of vocabulary richness; the former measures the proportion of unique words to the total word count, while the latter tracks hapax legomena (words appearing only once). 
We utilized Shannon Entropy to quantify the unpredictability of word choice, where higher values reflect a more varied and less repetitive communicative style. 
The Gunning Fog Index and Flesch Reading Ease are used to estimate the formal education level required to comprehend the text. 
We analyze syntactic structure through (i) Average Tree Depth, which measures grammatical nesting, and (ii) Standard Deviation of Sentence Length.
To evaluate the quality of the stylized text against the gold standard, we calculate the BLEU~\cite{BleuPapineniRWZ02} score, providing a quantitative measure of n-gram overlap and linguistic fidelity between the model outputs and the reference datasets. 
The G-Vendi score~\cite{jung2026prismatic} has been shown to be an effective proxy metric for evaluating the utility of synthetic data in downstream tasks. 
We use the \roberta model to compute G-Vendi scores for the different synthetic datasets generated for the \privateData dataset.

\input{tables/statistics-datasets}
\subsection{Models \& Parameters} \label{sec:params}
We will explain the models used for each task in the following.
\begin{itemize}[leftmargin=*,nosep]
    \item Synthetic data generation and filtering: We use \texttt{gpt-4.1-mini} for data generation with \texttt{temperature=0.9}. For filtering we experiment with different versions of \texttt{gpt-4.1-mini}, \texttt{gpt-4.1}, and \texttt{gpt-5o-mini}. In the final results, we use the \texttt{gpt-4.1-mini} model with the following parameters: \texttt{temperature}=0.1 and \texttt{top\_p}=1. To generate sequences of intents, we use \texttt{gpt-4.1} with \texttt{temperature}=0.9 to encourage diversity.
    
    \item Stylization: We use \texttt{T5} and \texttt{Llama-3.2-1B} models for this task. For \texttt{T5} we train the model for seven epochs with \texttt{learning\_rate}=3e-4, \texttt{batch\_size}=8, \texttt{weight\_decay}=0.01, evaluating the model every 400 steps and picking the best model based on validation loss. We set the window size of input to 512 and 64 for  \exampleStylizer and \universalStylizer variations, respectively. 
    The output length is set to 128 tokens.
    We fine-tune the \texttt{Llama-3.2-1B} model for four epochs with q-LoRA~\cite{LoraHuSWALWWC22} technique using q-LoRA parameters of \texttt{r}=16, \texttt{alpha}=16, and \texttt{dropout}=0.05. We use BLEU~\cite{BleuPapineniRWZ02} metric to evaluate the validation set every 100 steps to select the best training checkpoint.
    The batch size of 32 with the same sequence length as T5 is used for training. \looseness=-1
    
    \item Intent classification: We do intent classification with \texttt{Llama-3.2-1B} and \texttt{distilroberta-base} models and select the best checkpoint based on the macro F1 metric of the validation set. 
    We repeat each experiment 5 and 3 times for \texttt{distilroberta-base} and \texttt{Llama-3.2-1B} using different seed values and report the average performance. 
    For \texttt{distilroberta-base} we train the model for 15 epochs, using the following parameters: \texttt{sequence\_length} = 512, \texttt{batch\_size} = 16, \texttt{learning\_rate} = 2e-5. 
   We fine-tune the \texttt{Llama-3.2-1B} model using q-LoRA for text generation over five epochs with the following parameters: \texttt{sequence\_length} = 512, \texttt{batch\_size} = 2, \texttt{learning\_rate} = 5e-5, and q-LoRA parameters: \texttt{alpha}=16,
    \texttt{dropout}=0.1, \texttt{r}=16.
\end{itemize}

\subsection{Prompts}\label{sec:app-prompts}
We explain the prompts designed for different parts of our methodology in this section.
\begin{itemize}[leftmargin=*,nosep]
    \item The designed prompt for the synthetic dialogue generation in Equation~\ref{eq:dialog_gen}, is shown in Table~\ref{tbl:prompts-dialog_gen}.
    The prompt continues the given dialogue by generating multiple turns with the given intent. We pass the user writing style, dialogue history, user travel information (i.e., topic attributes) with the intent definition to the LLM.
    \item An example of a class definition for SGD dataset is shown in Table~\ref{tbl:class-def}. As can be seen, the class definition specifies the user’s objective in the generated dialogue, along with the minimum and maximum length of the dialogue chunk and the information required by the system. The prompt defines the system’s behavior by clearly indicating which information must be obtained before executing the user’s requested task. Consequently, if the necessary information is not provided in the context, the system may ask clarifying or elicitation questions to gather the missing details.
    \item The prompt used for generating different possible sequences of intents for dialogue generation is shown in Table~\ref{tbl:prompts-sequence-gen}. The prompt includes a set of intents along with a list of requirements that specify how different intent classes typically relate to one another. It instructs the LLM to generate $N$ sequences. Additionally, it allows an optional list of intents to be provided, in which case the model is guided to generate sequences that each include at least one of the specified intents.
    \item The prompt designed for generating different possible values for the attribute dimensions is shown in Table~\ref{tbl:prompts-attr-gen}. We only pass the attribute dimension as input and instruct the LLM to generate different possible values for the given dimension.
    \item The prompt designed for Filtering is shown in Table~\ref{tbl:prompts-filtering}. We put definition of each class and one example per each class in the prompt. We pass the current intent to LLM and ask the LLM to predict the intent and check if it is same as the given intent or not.
    \item The prompt designed for generating training data for stylization is shown in Table~\ref{tbl:prompts-style-data-gen}. 
    \item We also put the prompt used for \universalStylizer and \exampleStylizer in Tables~\ref{tbl:prompts-style-univ} and~\ref{tbl:prompts-style-examp}, respectively.
\end{itemize}
Some example attributes and values for the SGD dataset are shown in Table~\ref{tab:attribute-values}.

\input{tables/prompts-dialog_gen}

\input{tables/prompts-sequence-gen}

\input{tables/class-def}
\input{tables/attributes-table}

\input{tables/prompts-attr_gen}

\input{tables/prompts-filtering}

\input{tables/prompts-stylization}

%% file: tables/statistics-datasets.tex
\begin{table}[]
    \centering
    \small
    \caption{Statistics of the intent classification datasets.}
    \resizebox{\columnwidth}{!}{
    \begin{tabular}{lcccc}
    \toprule
     \multirow{2}{*}{\textbf{Dataset}}  & \multirow{2}{*}{\textbf{Num. intents}} &      &  \textbf{Size}  &  \\ \cmidrule{3-5}
                  &     & \textbf{Train} & \textbf{Test}  & \textbf{Eval} \\ \toprule
    \privateData  & 11  & 7,290          & 1,860          & 1,779 \\
    SGD           & 19  & 117,662      & 28,566    & 19,302 \\
    \bottomrule              
    \end{tabular}
    }
    \label{tab:dataset-stat}
\end{table}

%% file: tables/prompts-dialog_gen.tex
\begin{table}[!h]
\small
\vspace{1em}
\caption{The prompt designed for synthetic dialogue generation (Equation~\ref{eq:dialog_gen}).}
\begin{tabularx}{\linewidth}{X}
\toprule
\#\textbf{ Instruction:} Imagine you are a user chatting with a chatbot. In this chatbot, the intent of the user is detected at each turn to call the appropriate agent. 
You will be given existing chat, some information about the user and writing style of user but some details might be unclear initially and will be clarified as the conversation progresses.
Your task is to continue the conversation by generating a chunk of chat including user utterance and system response with the given intent. 
The generated conversation must be about the given intent.\\
\addlinespace
\#\textbf{ Intent:} {$c_i$}\\
\#\textbf{ The user's writing style}: {$w$} \\
\#\textbf{ Given Travel Information:} {$t$} \\
\#\textbf{ Chat:} {$D^{(i-1)}$} \\
\addlinespace
\textbf{\# Output format JSON:} \\
\texttt{[ \{"Human": , "AI": \}, \{"Human": , "AI": \},..]} \\
\bottomrule
\end{tabularx}
\label{tbl:prompts-dialog_gen}
\end{table}

%% file: tables/prompts-sequence-gen.tex
\begin{table}[!h]
\small
\vspace{1em}
\caption{The prompt designed for generating possible sequences of intent. The sequences of intents will be used as input in the dialogue generation function. The intent classes in this example are from the SGD dataset.}
\begin{tabularx}{\linewidth}{X}
\toprule
\#\textbf{ Instruction:} You are simulating a user interacting with a travel planning chatbot. Here is the list of possible intents: \\
FindBus: Find a bus itinerary between cities for a given date \\
BuyBusTicket: Buy tickets for a bus itinerary \\
SearchOnewayFlight: Search for a one-way flight with your set of preferences \\
SearchRoundtripFlights: Search for round-trip flights with your set of preferences \\
GetCarsAvailable: See available cars for rental in a particular city and a date \\
ReserveCar: Reserve a rental car for the specified pickup location and dates \\
...\\
\addlinespace
\#\textbf{ Requirements:} \\
1. Usually, ReserveHotel intent comes after SearchHotel intent in a conversation.\\
2. Usually, ReserveCar intent comes after GetCarsAvailable intent in a conversation. \\
3. Usually, ReserveRestaurant intent comes after FindRestaurants intent in a conversation. \\
4. Usually, the BuyBusTicket intent comes after the FindBus intent in a conversation. \\
5. Usually, the BookAppointment intent comes after the FindProvider intent in a conversation. \\
5. Usually, the BuyEventTickets intent comes after the FindEvents intent in a conversation. \\
\addlinespace
\# The generated sequences should include at least one of the following intents \{list of intents\} and one of the following intents \{list of intents\}.\\
Please generate \{N\} realistic sequences of intents, representing the order in which a user might express these intents in a single conversation. 
The generated sequences must be diverse and realistic. Do not generate repetitive sequences. Each sequence should include 1--4 intents
Output only the sequences as lists of intent names. \\
\addlinespace
\bottomrule
\end{tabularx}
\label{tbl:prompts-sequence-gen}
\end{table}

%% file: tables/class-def.tex
\begin{table}[!h]
\small
\vspace{1em}
\caption{An example of class definition. Definition of ReserveHotel intent class in SGD dataset.}
\begin{tabularx}{\linewidth}{X}
\toprule
Continue the given chat with a short conversation (1-5 turns) between a user and a system.\\
The user wants to reserve a hotel. \\
\addlinespace
\# \textbf{ Requirements:}\\
1. The system should know the name of the hotel, name of the city, check-in date and number of days to stay before reserving the hotel.\\
2. The system should ask clarification or elicitation questions to get the required information if they are not mentioned in the chat history. The system must never produce acknowledgment-only, confirmation, or ``working on it'' responses.\\
3. The system's language is friendly and supportive, offering polite clarification and gentle questions to gather details. It uses short sentences and avoids too much details and long response.\\
4. The conversation concludes when the system reserves the hotels and confirms it.\\
\addlinespace
\# \textbf{ Note:} Please do not generate more than 5 turns of conversation.\\
\addlinespace
\bottomrule
\end{tabularx}
\label{tbl:class-def}
\end{table}

%% file: tables/attributes-table.tex
\begin{table}[]
\tiny
    \centering
    \caption{Examples of attribute dimensions and values defined for SGD dataset.}
    \begin{tabular}{llp{4cm}}
    \toprule
Attribute Type & Dimension & Values  \\ \midrule
\multirow{2}{*}{Class-independent}  &  Number of people    &  ``1 person'', ``2 people'', ``3 people'', ``a couple with 1 child'', ``a couple with teenagers'', ``family of 6``       \\ \cmidrule{2-3}
     &  Location    &    ``from Belgium, Brussels to  Austria, Vienna'', ``from South Africa, Cape Town to India, Mumbai''    \\ \midrule

\multirow{7}{*}{Class-dependent}     &  Restaurant    &  ``with vegetarian options'', ``vegan menu available'', ``gluten-free options'', ``kosher restaurant''       \\ \cmidrule{2-3}
     &  Car rental    &   ``a hybrid car with mid-price'',``an automatic SUV with full coverage insurance'', ``a luxury sedan with premium budget'', ``a manual hatchback with basic insurance'',      \\ \cmidrule{2-3}
     &  House    &    ``a furnished apartment with two bedrooms'', ``a house with three bathrooms and a garden'', ``an unfurnished studio under \$1000 per month'',      \\ \cmidrule{2-3}
     &   Flight   &  ``an economy class ticket with window seat'', ``a business class flight with vegetarian meal'', ``a direct flight with aisle seat preference'', ``a first class ticket with extra baggage'',       \\     \cmidrule{2-3}
     & Bus & ``a sleeper bus with window seat'', ``an AC bus with snacks included'', ``a standard bus under \$20'', ``a luxury bus with WiFi and charging port`` \\  \cmidrule{2-3}
     & Hotel   &  ``a hotel with a good view'', ``a hotel with two queen beds'', ``a hotel with free breakfast'', ``a hotel with a swimming pool'', ``a hotel with a family suite'', ``a hotel with a kitchenette``\\  \cmidrule{2-3}
     &Movie &  ``a comedy movie in English released in 2023'', ``an action film from the USA with a PG-13 rating'', ``a French drama from the 1990s available on Netflix`` \\
     
\bottomrule      
    \end{tabular}
    \label{tab:attribute-values}
\end{table}

%% file: tables/prompts-attr_gen.tex
\begin{table}[!h]
\small
\vspace{1em}
\caption{The prompt designed for attribute value generation.}
\begin{tabularx}{\linewidth}{X}
\toprule
\#\textbf{ Instruction:} I want to generate synthetic data for a specific class of intent in trip planner chatbot. To this aim, I want to generate different possible values for the given attribute dimensions. Consider the attribute of \{attribute\_name\}, think of different values that it can have. Please generate attribute values as much as possible. \\
\addlinespace
\#\textbf{ Attribute values:}\\
\addlinespace
\bottomrule
\end{tabularx}
\label{tbl:prompts-attr-gen}
\end{table}

%% file: tables/prompts-filtering.tex
\begin{table}[!h]
\small
\vspace{1em}
\caption{The prompt designed for Filtering.}
\begin{tabularx}{\linewidth}{X}
\toprule
\#\textbf{ Instruction:} You are given a list with intents and their definition, a chat between a user and an AI assistant, and the intent predicted by an intent detection model for the last user message. \\
Your task is to act as a judge and determine whether the intent detection model has predicted the right intent or not and, if not, to suggest what the right intent is.\\
To identify the most relevant intent, you need to consider the last user utterance and the history of the chat.\\
Sometimes the user utterance does not express any of the intents described. In that case, the chatbot decides not to call any agent. This case is called "other" intent.\\
1. Carefully read the intent descriptions and the chat.\\
2. If the user is simply answering a system question that is meant to clarify or elicit more information about their original request, the intent remains the same as the original request.\\
3. Decide if the last user utterance expresses the intent predicted by the intent detection model or another intent.\\
4. Tell your reasoning in the response. Keep the reasoning short. \\
\addlinespace

\textbf{\# Intents descriptions:} \\
1. FindMovies: user wants to find movies by genre and optionally director, or search for movies by location, genre or other attributes.\\
2. GetWeather: user wants to get the weather of a certain location on a date.\\
...\\
\addlinespace

\textbf{\# Example 1:}\\
\textbf{\# Chat:}\\
"system: Should I reserve a table for you in Thai House \& Wine Bar?" \\
"user: Yes, please make a reservation for morning 11:45." \\
\textbf{\# Intent detection model prediction:} "ReserveRestaurant" \\
\textbf{\# Output:}\\
\{ "reason": "user wants to reserve a table in a restaurant.", \\
  "is\_prediction\_correct": "yes",\\
  "your\_prediction": "ReserveRestaurant"\} \\
\addlinespace
\#\#\#\#\#\#\#\#\#\#\#\# \\
......... \\
\#\#\#\#\#\#\#\#\#\#\#\# \\
\addlinespace

\textbf{\# Example n:} \\
\textbf{\# Chat:} \{chat\} \\

\textbf{\# Intent detection model prediction:} \{given\_intent\} \\
\textbf{\# Output:} \\
\{ \\
"reason": [your\_reason], \\
"is\_prediction\_correct": "Yes" or "No", \\
"your\_prediction": [one of the intent classes] \\
\} \\
\bottomrule
\end{tabularx}
\label{tbl:prompts-filtering}
\end{table}

%% file: tables/prompts-stylization.tex
\begin{table}[!h]
\small
\vspace{1em}
\caption{The prompt designed for generating training data of stylization models.}
\begin{tabularx}{\linewidth}{X}
\toprule
\#\textbf{ Instruction:} I will give you a conversation between a user and an AI system. Your task is to rewrite the last user utterance in a more detailed and clear way in your own language.\\
The rewritten sentence must preserve the original meaning and intent of the user’s utterance and must not introduce any new information that is not already implied or stated in the conversation.\\
Please rewrite only the last "user" utterance five times.\\
\addlinespace
\#\textbf{ Conversation:} \{conversation\} \\
\addlinespace
Rewrite of the last user utterance: \\
\addlinespace
Generate output in this JSON format:\\
\addlinespace
\{"rewrite": \{'sent1', 'sent2', ..., 'sent5'\} \} \\
\addlinespace
\bottomrule
\end{tabularx}
\label{tbl:prompts-style-data-gen}
\end{table}

\begin{table}[!h]
\small
\vspace{1em}
\caption{The prompt designed for the \universalStylizer stylization model.}
\begin{tabularx}{\linewidth}{X}
\toprule
\#\textbf{ Instruction:} Rewrite the last LLM utterance in the style and language of the 'Human' from the given conversation, preserving the original meaning and intent.\\
\addlinespace
\#\textbf{ Input:} {}\\
\#\textbf{ Output:} {}\\
\addlinespace
\bottomrule
\end{tabularx}
\label{tbl:prompts-style-examp}
\end{table}

\begin{table}[!h]
\small
\vspace{1em}
\caption{The prompt designed for the \exampleStylizer stylization model.}
\begin{tabularx}{\linewidth}{X}
\toprule
\#\textbf{ Instruction:} Rewrite the LLM utterance in the style and language of a real human, preserving the original meaning and intent.\\
\addlinespace
\#\textbf{ Input:} {}\\
\#\textbf{ Output:} {}\\
\addlinespace
\bottomrule
\end{tabularx}
\label{tbl:prompts-style-univ}
\end{table}